\documentclass[lettersize,journal]{IEEEtran}
\usepackage{amsmath,amsfonts}
\usepackage{algorithmic}
\usepackage{algorithm}
\usepackage{array}
\usepackage[caption=false,font=footnotesize,labelfont=rm,textfont=rm]{subfig}
\usepackage{textcomp}
\usepackage{stfloats}
\usepackage{url}
\usepackage{verbatim}
\usepackage{graphicx}
\usepackage{cite}
\usepackage[T1]{fontenc}
\usepackage{booktabs}
\usepackage{multirow}
\usepackage{footnote}
\usepackage{threeparttable}

\usepackage{color}
\definecolor{myColor}{RGB}{0,0,200}

\begin{document}
	
	\title{Global-Local MAV Detection under Challenging Conditions based on Appearance and Motion}
	
	\author{Hanqing Guo, Ye Zheng, Yin Zhang, Zhi Gao, Shiyu Zhao\vspace{-2em}
		\thanks{H. Guo is with the College of Computer Science and Technology at Zhejiang University and the School of Engineering at Westlake University, Hangzhou, China. (e-mail: guohanqing@westlake.edu.cn)}
		\thanks{Y. Zheng, Y. Zhang, and S. Zhao are with the School of Engineering at Westlake University, Hangzhou, China. (e-mail: zhengye@westlake.edu.cn; zhangyin@westlake.edu.cn; zhaoshiyu@westlake.edu.cn)}
		\thanks{Z. Gao is with the School of Remote Sensing and Information Engineering, Wuhan University, Wuhan, China. (e-mail: gaozhinus@gmail.com)}
		\thanks{(\textit{Corresponding author: Shiyu Zhao.})}
	}
	
	\maketitle
	
	\begin{abstract}
		Visual detection of micro aerial vehicles (MAVs) has received increasing research attention in recent years due to its importance in many applications. However, the existing approaches based on either appearance or motion features of MAVs still face challenges when the background is complex, the MAV target is small, or the computation resource is limited.
		In this paper, we propose a global-local MAV detector that can fuse both motion and appearance features for MAV detection under challenging conditions. This detector first searches MAV targets using a global detector and then switches to a local detector which works in an adaptive search region to enhance accuracy and efficiency. Additionally, a detector switcher is applied to coordinate the global and local detectors. A new dataset is created to train and verify the effectiveness of the proposed detector. This dataset contains more challenging scenarios that can occur in practice. Extensive experiments on three challenging datasets show that the proposed detector outperforms the state-of-the-art ones in terms of detection accuracy and computational efficiency. In particular, this detector can run with near real-time frame rate on NVIDIA Jetson NX Xavier, which demonstrates the usefulness of our approach for real-world applications. The dataset is available at \emph{https://github.com/WestlakeIntelligentRobotics/GLAD}. In addition, A video summarizing this work is available at \emph{https://youtu.be/Tv473mAzHbU}.
	\end{abstract}
	
	
	\section{Introduction}
\begin{figure}[t]
		\centering
		\subfloat[Complex background.]{\includegraphics[width=0.98\linewidth]{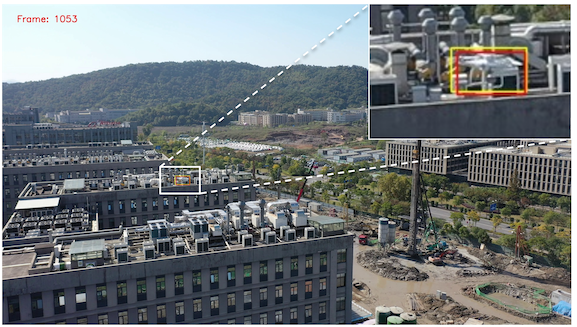}}\vspace{1mm}
		\subfloat[Small MAV (8x8 pixels).] {\includegraphics[width=0.98\linewidth]{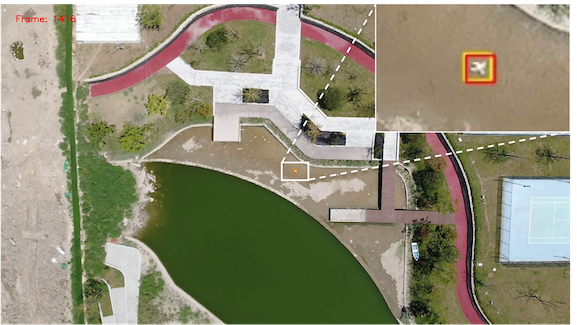}}
		\caption{Examples of challenging conditions for MAV detection. Our approach can work effectively in very challenging conditions such as complex backgrounds and small-sized MAVs. \textbf{Red box} indicates ground-truth. \textbf{Yellow box} indicates the predicted bounding box by the local detector of our approach. (Top right images are better view with 500$\%$ zoom-in)}
		\label{fig_1}
	\end{figure}

	Vision-based MAV detection has attracted increasing attention in recent years due to its application in many tasks such as vision-based swarming\cite{tang2018vision, 2018TM, 2020marker}, aerial see-and-avoid\cite{Sapkota2016Vision, 2021sense-avoid}, and malicious MAV detection\cite{zhang2018survey, 2019Unlu, 2020realworld}. Different from the existing works\cite{2022Camera, Xie2021SmallLT, 2020DroneStaticCamera} that consider the ground-to-air scenario, this work focuses on the air-to-air scenario where a camera carried by a flying MAV is used to detect another flying target MAV. The air-to-air scenario is more challenging than the ground-to-air scenario because the camera itself is moving and the target MAV is often engulfed by complex background scenes such as buildings, trees, and other man-made things when the camera looks down from the top (Fig.~\ref{fig_1}(a)). Moreover, the target MAV may be extremely small in the image when seen from a distance (Fig.~\ref{fig_1}(b)). In addition, the detection algorithm must be sufficiently efficient in order to be implemented onboard with limited computational resources.
	
	In recent years, many appearance-based methods that rely on deep learning techniques have been proposed for vision-based MAV detection. For example, some state-of-the-art object detection networks such as YOLO series, R-CNN series, SSD, DETR \cite{2021DT-Benchmark, 2021Air, 2022Anti-UAV-DT} have been applied to MAV detection. These methods usually work effectively in relatively simple scenarios where the target MAV is distinct from the background and its size is relatively large in the image. However, appearance features are not stable in more complex scenarios. For example, as shown in Fig.~\ref{fig_1}(a), the background scene is extremely complex when the camera looks down from the top. The target MAV can easily get engulfed in complex background scenes such as trees or buildings.  Moreover, when the target MAV flies far away from the camera, its image may only occupy a tiny portion of the image. For example, as shown in Fig.~\ref{fig_1}(b), an MAV seen from nearly 100~m only occupies 8$\times$8 pixels in an image of 1920$\times$1080 pixels. Although we can zoom in to detect distant MAVs, it would lead to a smaller field of view which is unfavorable to search and track moving MAVs. It is therefore necessary to develop high-performance algorithms to detect MAVs under challenging conditions.
	
	Motion features are helpful in detecting MAVs under challenging conditions. Many motion-assisted methods which combine appearance features and motion features have been proposed to detect MAVs based on, for example, background subtraction\cite{2021Fast, NPU2020}, low-rank based methods\cite{2017UDT, wang2019flying}, spatio-temporal information\cite{2022Camera, Xie2021SmallLT, Xie2020AdaptiveSS}, and optical flow \cite{Rozantsev2017DetectingFO, 2021Dogfight, wang2023RAFT}. However, motion-assisted MAV detection still faces the following challenges. First, motion cues of MAVs are difficult to separate from the background when the background is non-planar or the camera moves drastically. In particular, most of the existing motion-assisted methods assume the background scene is planar \cite{2021Fast, 2020hybridICUAS, zhang2021jointly} and use affine transformation or perspective transformation to estimate the camera movement. Although this assumption is valid in some cases especially when the camera flies at a high altitude so that the height of the ground object is neglectable, it is still invalid when the flight altitude is relatively low. In this case, the ground objects such as buildings, trees, and lampposts may violate the planar assumption. Second, many existing motion-assisted methods such as the region-based sliding windows in \cite{Rozantsev2017DetectingFO} or the two-stage approach in \cite{2021Dogfight} are computationally intensive and hence difficult to implement in onboard computers of MAVs. Third, most of the existing motion-assisted methods can only detect moving MAVs. Since multirotor MAVs may hover stationarily, it is necessary to develop a method that can detect both stationary and moving MAVs by integrating different types of features.
	
	To overcome the limitations of the existing approaches, we propose a new \underline{G}lobal-\underline{L}ocal M\underline{A}V \underline{D}etector (GLAD) for air-to-air detection of MAVs under challenging conditions. This algorithm is composed of a global detector, a local detector, an adaptive search region, and a detector switcher. First, the global detector is used to search for MAV targets in the full-size image. It consists of a global appearance-based detection module (GAD) and a global motion-based detection module (GMD). The GMD serves as a good assistant when the appearance features are unreliable. Second, after the MAV target has been detected by the global detector, the local detector is activated to conduct subsequent detection in a Kalman filter-based adaptive search region cropped from the neighboring area around the target. This local detector consists of a local appearance-based detection module (LAD) and a local motion-based detection module (LMD) as well. It can significantly improve the detection accuracy under challenging conditions because the target's resolution in the adaptive search region is greatly improved compared with the method based on down-sampling. Third, a detector switcher is designed to adaptively coordinate the global and local detectors. The detector switcher can adaptively switch between the global and local detectors based on the detection results of the previous frames. It can avoid the local detector searching in the wrong search region when the local detector fails for a while.
	
	The main contributions and novelties of this work can be summarized as follows.
	
	1) We propose a global-to-local method that can significantly improve the accuracy and efficiency for MAV detection under challenging conditions. This has been verified by experimental results on three challenging datasets, showing that our proposed method outperforms the existing ones including \cite{mega, 2021Dogfight, TPH-YOLOv5}.
	
	2) We design a motion-based classifier and an appearance-based classifier that can effectively and efficiently eliminate the interruptions generated by imperfect image alignment in non-planar scenes. This is supported by experimental results in non-planar scenes where 3D structures such as high buildings, trees, and lampposts are dominant rather than a planar background assumption which is commonly used in aerial object detection such as \cite{2021Fast, 2020hybridICUAS, zhang2021jointly}.
	
	3) We introduce a Kalman filter-based adaptive search region to dynamically adjust the size and location of the local search region. The experimental results show that this method can further enhance the robustness of our method in the presence of occlusion and missing detection.
	
	4) We create a new dataset, named ARD-MAV, which contains 60 videos and 106,665 frames. This dataset contains more challenging scenarios that may occur in practice. Compared to the existing datasets \cite{2021Fast, Rozantsev2017DetectingFO, 2022Anti-UAV-DT, Drone-vs-Bird}, our proposed dataset has the smallest average object size. It contains various real-world challenges such as complex backgrounds, 3D structures, abrupt camera movement, and small MAVs.
	
	\section{Related Works}\label{section_relatedWork}
	\subsection{Appearance-Based MAV Detection} The existing appearance-based MAV detection works can be classified into conventional methods and deep learning methods. The conventional methods usually use feature extraction methods to obtain target features and then use a discriminant classifier to classify the MAV. In particular, the work in \cite{2015Vision} tests Harr-like features, histogram of gradients (HOG), and local binary patterns (LBP) using cascades of boosted classifiers for MAV detection. The work in \cite{Sapkota2016Vision} uses the Adaboost algorithm with HOG features for online detection of MAV. The work in \cite{2018Unlu} uses 2-dimensional, rotation, and translation invariant Generic Fourier Descriptor (GFD) features and classifies targets as a drone or bird by a neural network. Besides, template matching and morphological filtering\cite{2018TM} have also been considered for MAV detection. These methods have been verified to be effective in simple cases. Nevertheless, when the shape and size of the MAV change vastly or the background is too complex, conventional methods usually have difficulties in these scenarios.
	
	On the other hand, with the fast development of deep learning in object detection, there emerge many works of MAV detection using deep learning methods. The work in \cite{2021Air} evaluates eight state-of-the-art deep learning algorithms on the Det-Fly dataset for MAV detection. Similarly, the authors in \cite{2021DT-Benchmark} evaluates four state-of-the-art deep learning algorithms on three representative MAV dataset (MAV-VID, Drone-vs-Bird, Anti-UAV). To further improve the object detection accuracy, the authors in \cite{2021PruneYOLOv4} implement a special augmentation method and prune the convolution channel and shortcut layer of YOLOv4 for small drone detection, the authors in \cite{rui2021comprehensive} propose a novel comprehensive approach that combines transfer learning based on simulation data and adaptive fusion to improve small object detection performance. Although deep learning methods have made great progress compared with conventional methods, there are still many challenges for appearance-based MAV detection such as complex backgrounds, motion blur, and small objects.
	
	\subsection{Motion-Assisted MAV Detection} Motion-assisted MAV detection methods aim at detecting the MAV by combining motion features and appearance features. The existing motion-assisted MAV detection methods can be classified into stationary cameras and moving cameras. The works in \cite{2017DeepCross-domain, 2020DroneStaticCamera} monitor the sky with a stationary camera and then use background subtraction and CNN-based object classification for MAV detection. The works in \cite{Xie2020AdaptiveSS, Xie2021SmallLT} fuse the spatiotemporal feature of the target for remote flying drone detection. The work in \cite{2022Camera} adopts appearance features to exclude non-MAV moving targets and then uses a motion-based classification algorithm to distinguish MAV from other interruptions.
	
	Compared with a stationary camera, MAV detection from a moving camera is much more challenging since the movement of the background is coupled with the movement of targets. The works in \cite{Li2016MultitargetDA, 2021Fast} propose a UAV-to-UAV video dataset and a general architecture for small MAV detection from a camera mounted on a moving MAV platform. The authors detect the moving MAV by subtracting neighboring frames and then identify MAV using a hybrid classifier. Similarly, the works in \cite{Rozantsev2017DetectingFO, 2015flying} create a more challenging dataset for detecting flying objects using a single moving camera. The authors first employ two CNN networks in a sliding window fashion to obtain the motion-stabilized spatial-temporal cubes and then use the third CNN network to classify MAV in each spatial-temporal cube. The work in \cite{2021Dogfight} proposes a two-stage segmentation approach. In the first stage, the authors utilize a 2-D convolution network and channel-pixel-wise attention to extract contextual information based on overlapping patches. Then, a 3-D convolution network and channel-pixel-wise attention are used to learn spatiotemporal cues and discover the missing detections of stage-1. The work in \cite{wang2023RAFT} proposes a feature super-resolution-based UAV detector with motion information extraction based on dense optical flow. However, these methods are either too time-consuming or only effective when the target is large enough or the background is very simple, which is challenging for air-to-air MAV detection in a cluttered environment.
	
	\section{An Overview of the Proposed Method}\label{section_overview}
\begin{figure*}[!t]
		\centering
		\includegraphics[width=0.99\linewidth]{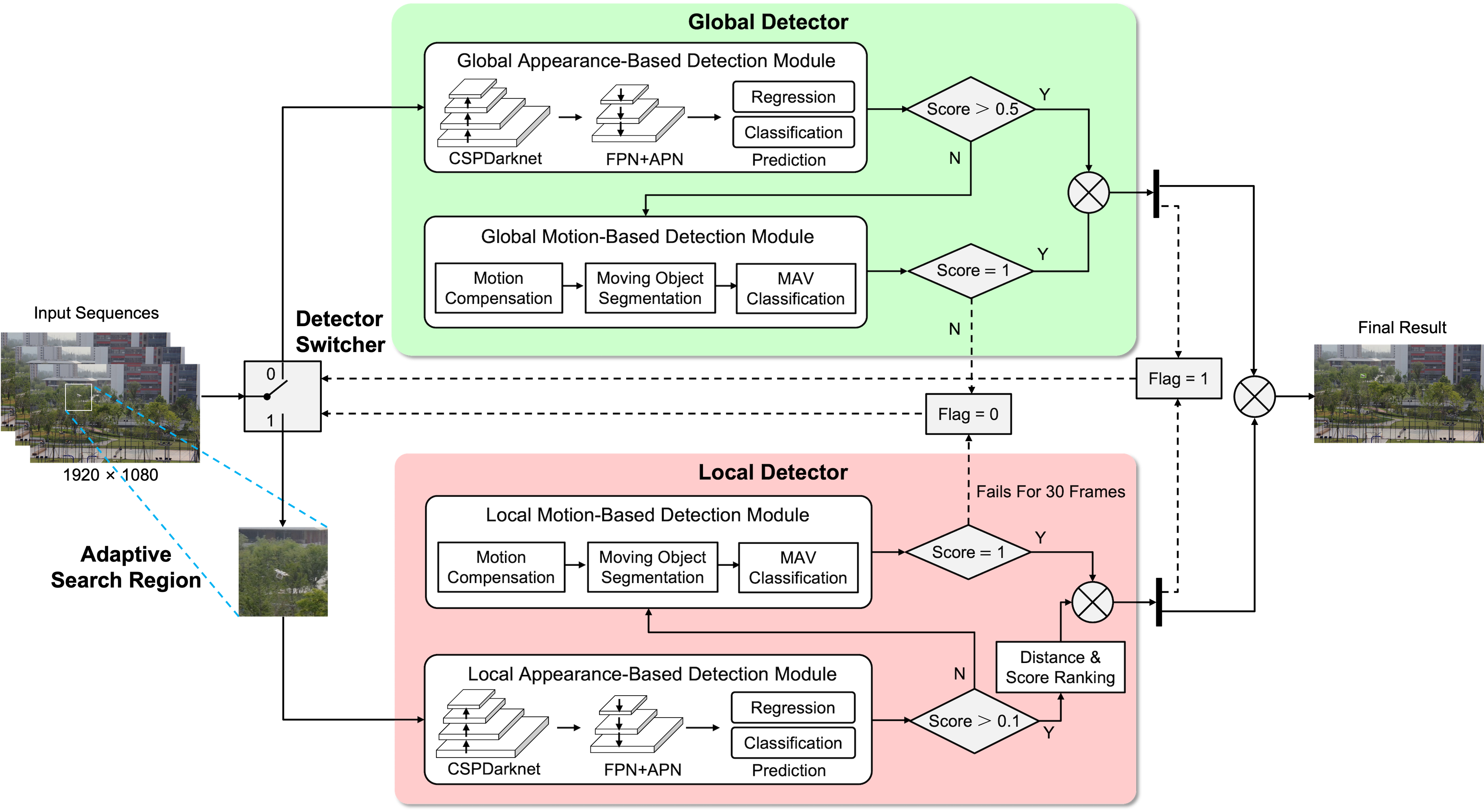}
		\caption{The architecture of our proposed GLAD algorithm. Given an input image of 1920$\times$1080, we first use the global detector to obtain the initial position of the target MAV. Then, a Kalman filter-based adaptive search region is cropped around the center of the last target as the local search region. The local detector is applied to the local search region for subsequent MAV detection. Finally, a detector switcher is applied to adaptively switch between the global and local detectors in order to avoid the local detector falling into the wrong search region. The solid arrow line represents the data flow. The dashed arrow line represents signal flow.}
		\label{fig_2}
	\end{figure*}

	To effectively detect MAVs under challenging conditions, we propose a global-local MAV detector called GLAD. The architecture of GLAD is illustrated in Fig.~\ref{fig_2}. It consists of a global detector, a local detector, an adaptive search region, and a detector switcher.
	
	\textbf{Global Detector:} The global detector is composed of an appearance-based module and a motion-based module.  First, YOLO is used as an appearance-based detection module for MAV detection in the full-size image. YOLO is a fast single-shot object detector based on convolutional neural networks. Our previous study \cite{2021Air} shows that YOLO can achieve a good balance between accuracy and speed.
	In this paper, YOLOv5s is trained with the proposed ARD-MAV dataset. Detailed information about the dataset is given in Section~\ref{dataset}.To reduce false alarms as many as possible, a high confidence threshold $T_0$ is adopted.
	
	When the appearance-based module fails to detect an MAV under challenging conditions, the motion-based module is activated. The motion-based module aims to detect a moving MAV and is composed of motion compensation, moving object segmentation, and MAV classification. Due to the extra temporal cues, the motion-based module shows superior performance under challenging conditions. The procedure of the motion-based detection module is presented in Section~\ref{motion-detector}. Since the motion-based module is more time-consuming, it will be called only when the appearance-based module fails. The appearance-based module and the motion-based module are continuously executed until an MAV is detected.
	
	\textbf{Local Detector:} After the target MAV has been found by the global detector, a local detector is activated to conduct subsequent MAV detection. Considering that the target usually does not move too far between two consecutive frames, a small area around the center of the target is cropped as the search region for MAV detection. Within the local search region, a local detector is applied to detect the MAV.
	
	The local detector is composed of an appearance-based module and a local motion-based module as well. Different from the global detector, the local detector is trained with cropped images and has different parameters settings. Since the local search region only focuses on a small patch of the full-size image, the resolution of the target can be greatly improved compared to the down-sampling method, and many non-MAV interruptions are removed as well. We set a low confidence threshold $T_1$ to detect as many targets as possible. Although a low threshold would bring some false targets, confidence ranking and distance ranking are applied to obtain the most reliable target. Specifically, the target with the highest confidence score or the closest distance to the last target is selected as the final target.
	
	The local detector not only improves the detection accuracy but also greatly improves the computational efficiency. Since the size of the local search region is much smaller than the size of the original image, the operations including bilinear interpolation, key points selection, optical flow matching, frame difference, and object classification in the local detector are much more efficient. The details about the computational efficiency are introduced in Section~\ref{inference}.
	
	\textbf{Adaptive Search Region:} Considering that most of the targets are small, a fixed size of 300$\times$300 area around the center of the target in the current frame is usually enough to cover the potential location of the target in the next frame. Nevertheless, when the detectors fail to detect a target for multiple frames, a fixed search region might not cover the area of the potential targets. To improve the robustness of our method when occlusion and missing detection happen, we design an adaptive search region to dynamically adjust the size and location of the local search region.
	
	To better predict the position of the local search region, we first use the Kalman filter to estimate the target position in the next frame. In particular, we use a Kalman filter to track the relative velocity of the target rather than tracking the target's position. We do this because the position is highly non-linear when the camera moves fast, however, the relative velocity is continuous in this scenario. The Kalman filter estimates the target state $x_t = (v_x, v_y, a_x, a_y)$ via a linear difference equation, 
		\begin{equation}
			x_t = M x_{t-1}  + w_{t-1},
		\end{equation}
		where $(v_x, v_y)$ denotes the velocity of the target center, $(a_x, a_y)$ denotes the acceleration of the target center in X-axis and Y-axis direction, $M \in \mathbb{R}^{4\times4}$ is the state transition matrix, $w_{t-1}$ denotes the modeling errors or process noise.
	
	In the prediction step, the target's velocity is obtained via a dynamic model expressed as, 
		\begin{equation}
			\hat {Z}_t = N x_t  + v_t,
		\end{equation}
		where $\hat{Z}_t$ denotes the predicted measurement, $N \in \mathbb{R}^{2\times4}$ is the measurement matrix, and $v_t$ is the measurement noise.
  
		In the updating step, the target state is updated with the actual measure $Z_t$ if the detection is successful. However, if the detection fails, the Kalman filter will not be updated, and we directly update the target state with the optimal prediction of the last frame.
	
	After we have obtained the estimated velocity, we estimate the target position in the current frame with the following equation, 
		\begin{equation}
			p_t = T(p_{t-1}; H)   + U_{t-1},
		\end{equation}
		where $p_t$ denotes the coordinate of the target center, $T$ and $H$ denotes the 2D perspective transformation and the homography matrix, $U_{t-1}$ is the velocity of the target center.
	
	Given the estimated target position, we set up a new search region in the next frame centered at the predicted target position. The size of the search region (a $L$ $\times$ $L$ square) is decided as following:
		\begin{equation}
			L = 300 + T_{lost} \times 4,
		\end{equation}
		where $T_{lost}$ is the number of frames that detectors fail to detect a target.
	
	\textbf{Detector Switcher:} Although the proposed local detector can significantly improve the detection accuracy and the adaptive search region can guide the local detector to the right search region in most cases, persistent detection failures may occasionally happen when the background is too complex or the target is too small to detect. Under these circumstances, the local detector may concentrate on a wrong search region where the true object is excluded. Besides, occlusion and abrupt camera movement may also cause the same problem. Hence, a detector switcher is designed to adaptively coordinate the global and local detectors.
	
	The switcher adaptively switches between the global detector and the local detector based on the detection results of the previous frames. Specifically, when the global and local detectors successfully detect a target MAV, the detector switcher will switch to or keep on the local detector in the next frame. When the global detector fails to detect the target MAV, the detector switcher will continue to execute detection using the global detector in the full-size image. However, if the local detector fails, the detector switcher will proceed to execute detection using the local detector in the next frame. The successive detection in the local search region can help the algorithm quickly recover from a failure status because the local detector has better detection accuracy and the target MAV usually does not move too far between consecutive frames. To avoid the local detector searching in the wrong search region, the detector switcher will switch to the global detector if the local detector fails for a while (for example 30 frames).
	
	\section{Motion-based Detection Module}\label{motion-detector}
\begin{figure*}[t]
		\centering
		\includegraphics[width=0.99\linewidth]{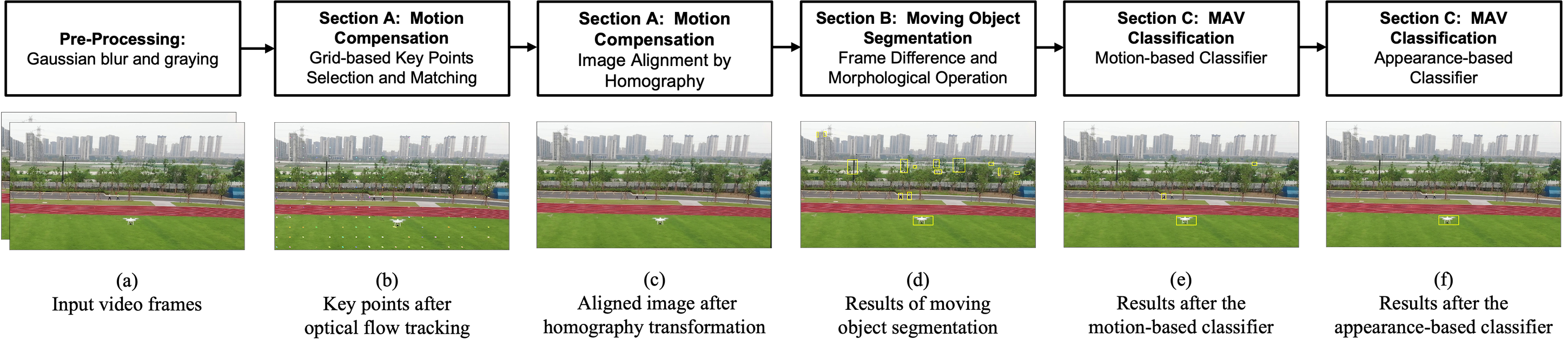}
		\caption{The flowchart of the motion-based detection module.}
		\label{fig_3}
	\end{figure*}

	This section introduces the procedure of the motion-based detection module in detail. The motion-based detection module is composed of three parts, namely motion compensation, moving object segmentation, and MAV classification (see Fig.~\ref{fig_3}). First, grid-based key points and perspective transformation are used to compensate for the camera motion. Second, frame difference and morphological operation are applied to segment moving objects. Third, a motion-based classifier and an appearance-based classifier are successively utilized to eliminate non-MAV moving objects and image alignment errors. The details are given as follows.
	
	\subsection{Motion Compensation}\label{motion-compensation}
	To separate the moving objects from the moving background, we must first align two adjacent frames so that the influence of the camera's ego-motion can be eliminated. In this paper, 2D perspective transformation is used for motion compensation because it exactly models the 2D background motion when the background results from the relative motion of a 2D plane in the 3D world and has been widely used for camera motion compensation in aerial view \cite{2021Fast, 2018effective, 2021uavdataset}. Perspective transformation requires the background to be planar or the camera only rotates. In many cases, this is a reasonable approximation since the pursuing MAV looks at the ground from a high altitude. Hence, most of the previous works only consider near planar scenes \cite{2021Fast, 2020hybridICUAS}. We noticed that non-planar scenes containing objects such as high buildings, lampposts, trees, and wire poles which are quite prevalent in the ARD-MAV dataset can remarkably influence the motion compensation quality and generate false positives. Nevertheless, our proposed motion-based classifier and the appearance-based classifier can remove these interruptions. 
	
	For computational efficiency and robustness in textureless regions such as sky and grassland, grid-based key points are used to calculate the homography matrix. We sample $30 \times 20$ key points uniformly distributed in each row and column across the previous frame. Then, these key points are tracked by the pyramidal Lucas-Kanade (LK) algorithm \cite{PLK} to obtain the corresponding points in the current frame (one example is shown in Fig.~\ref{fig_3}(b)). After the key points are matched over two adjacent frames, the homography matrix $H$ is calculated with the RANSAC method to reject outliers. The image in the previous frame $F_{n-1}$ can be aligned with the current frame $F_n$ by the perspective transformation. 
		\begin{equation}
			\hat {F}_{n-1}(x, y)= \emph{H} F_{n-1}(x, y).
		\end{equation}
		Here, $H$ represents the transformation matrix between $F_{n-1}$ and $F_n$, $ \hat {F}_{n-1}$ denotes the motion-compensated previous frame. One example of the motion-compensated frame is shown in Fig.~\ref{fig_3}(c).
	
	\subsection{Moving Object Segmentation}
	After we have obtained the motion-compensated previous frame, we can highlight the moving areas with absolute differences between the current frame and the motion-compensated previous frame. In this paper, frame difference and morphological operation are used for moving object segmentation. 
	
	\subsubsection{Frame Difference}The frame difference is defined as follows:
	\begin{equation}
		E_{n-1} = |I_n (x, y) - \hat {I}_{n-1} (x,y)|,
	\end{equation}
	where $I_n$ is the gray value of the current frame and $ \hat {I}_{n-1}$ is the gray value of the motion-compensated previous frame.
	
	Next, a threshold $T_2$ is applied on $E_{n-1}$ to remove noises and highlight the silhouette mask of the potential moving foreground. The pixel values above the threshold $T_2$ are set to 255 as foreground, and below the threshold $T_2$  are set to 0 as background. As a result, we obtain the binarized frame difference $D_{n-1}$. Concerning frame difference, the choice of threshold $T_2$ greatly influences the final result of moving object segmentation. If $T_2$ is too small, the noise will stand out and fill the image. Conversely, if $T_2$ is too big, some parts of the moving foreground will be wiped out. Moreover, a fixed $T_2$ can hardly adapt to the change of light intensity and moving background. Therefore, we correct the threshold $T_2$ with a light intensity correction term $T_A$ and a background motion correction term $T_M$.
	\begin{align}
		D_{n-1} &=
		\begin{cases}
			255, & { \text {if } } E_{n-1} > T_2 + \alpha T_A + \beta T_M \\
			{0,} &{\text{otherwise.}}
		\end{cases}  \\
		T_A &= \frac {1} {N_A}  \sum _ {(x, y) \in {A}} |I_n(x,y) - I_{n-1} (x,y)| \\
		T_M &= \frac {1} {N_S}  \sum _ {(x, y) \in {S}}  |P_n(x,y) - P_{n-1} (x,y)|
	\end{align}
	
	Here, $A$ represents the whole image pixels, $N_A$ represents the total number of pixels, and $\alpha$ is the light intensity suppress coefficient. $S$ denotes the set of matched key points, $N_S$ represents the total number of the rectified key points, $P$ is the pixel coordinates of the rectified key points, and $\beta$ is the background motion correction coefficient.
	
	\subsubsection{Post Processing} Frame difference method can segment the pixels belonging to moving objects, but may also contain noises, small holes, and disconnected blobs, which would cause wrong bounding boxes of the moving objects. Hence, multiple morphological operations are used to obtain the intact bounding boxes. Firstly, the morphological open and close operation are used iteratively on binarized frame difference to eliminate isolated pixels and fill the holes. Then, connected component analysis is used to obtain the total number of pixels of each object, and blobs whose area is below 30 pixels are eliminated because such small objects are usually difficult to recognize in the subsequent appearance-based classification. Finally, the maximum bounding rectangle is applied to mark the moving objects and obtain the bounding boxes. Among these bounding boxes, some adjacent bounding boxes that belong to the same object may be isolated from each other. Therefore, we merge the bounding boxes whose distance between another neighboring bounding box is smaller than $D_1$.
	
	\subsection{MAV Classification}
	Until now, we have obtained the candidate moving objects. However, there are still many non-MAV moving objects such as cars, pedestrians, swaying trees, shimmering water, and image alignment errors (see Fig.~\ref{fig_3}(d)). Hence, a motion-based classifier and an appearance-based classifier are successively used to separate MAV from these interruptions.
	
	\subsubsection{Motion-based Classifier} In principle, the motion features of MAVs are significantly different from interruptions such as swaying trees, shimmering water, and image alignment errors which are usually irregular in moving direction and moving amplitude. Therefore, most of these interruptions can be filtered by their statistical features. Assuming that MAV is a non-deformable object, and motion vectors between two consecutive frames are consistent. We firstly extract Shi-Tomasi corner points in $k^{th}$ moving object of frame $n$, then define the motion feature $f^{(k)}_n $ as the angle variance of motion vectors in Equation~\eqref{eq:motion_feature_equation}, and $g^{(k)}_n $ as the velocity variance of motion vectors in Equation~\eqref{eq:appearance_feature_equation}.
	\begin{align}
			f^{(k)}_n &= \frac{\sum\limits_{{d_t} \in {D^{(k)}_n}}({\arctan{{d_t}}} - \mu^{(k)}_n)^2}{S^{(k)}_n}
			\label{eq:motion_feature_equation} \\
			\mu^{(k)}_n &= \frac{\sum\limits_{{d_t} \in {D^{(k)}_n}}{\arctan{{d_t}}}}{S^{(k)}_n} \\
			g^{(k)}_n &= \frac{\sum\limits_{{d_t} \in {D^{(k)}_n}}(||{d_t}||- \lambda^{(k)}_n)^2}{S^{(k)}_n}
			\label{eq:appearance_feature_equation}\\
			\lambda^{(k)}_n& = \frac{\sum\limits_{{d_t} \in {D^{(k)}_n}} ||{d_t}||}{S^{(k)}_n}
	\end{align}
	Here, $D^{(k)}_n$ denotes the set of motion vectors of the $k^{th}$ moving object of frame $n$ and $S^{(k)}_n$ is the number of motion vectors. Given the above motion features, we build a motion-based classifier to eliminate these interruptions. We denote $y^{(k)}_n $ as a classification label where the zero value indicates that the $k^{th}$ object is noise, otherwise the $k^{th}$ object is the candidate MAV. The motion-based classifier is defined as follows:
	\begin{equation}
		y^{(k)}_n =
		\begin{cases}
			0, & { \text {if } } {f^{(k)}_n > T_3} \ or \ {{g^{(k)}_n} > T_4 \ or \ \lambda^{(k)}  < T_5}, \\
			{1,} &{\text{otherwise}}.
		\end{cases}
	\end{equation}
	where $T_3$ is the empirical threshold for angle variance, $T_4$ is the empirical threshold for velocity variance, and $T_5$ is the empirical threshold for velocity amplitude.
	
	\subsubsection{Appearance-based Classifier} 
\begin{figure}[h]
		\centering
		\includegraphics[width=0.9\linewidth]{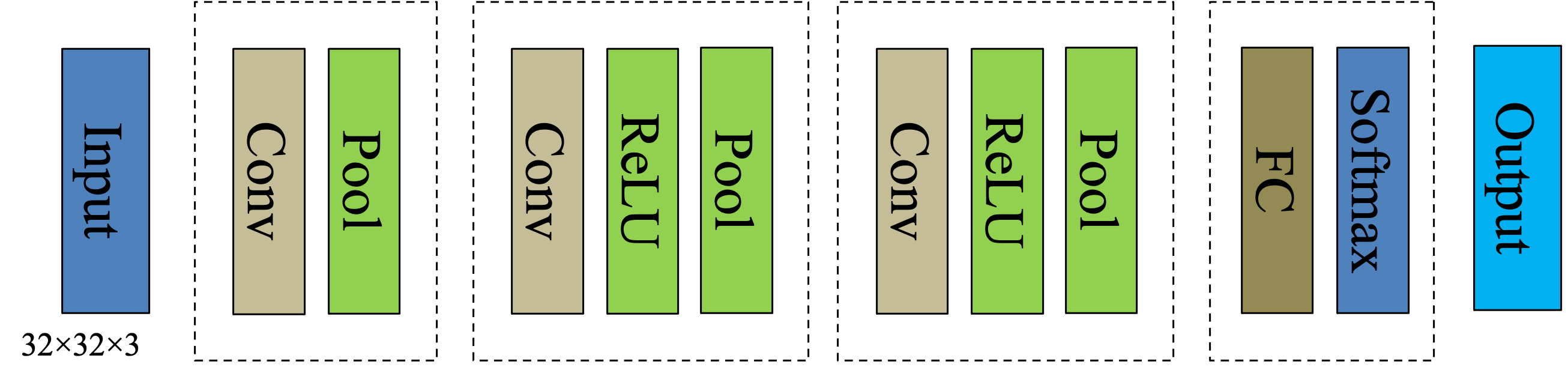}
		\caption{The architecture of the appearance-based classifier. The input to the network is a 32$\times$32$\times$3 image. The output represents the image is an MAV or a clutter.}
		\label{fig_6}
	\end{figure}

After the filtering of the motion-based classifier, the candidate moving objects still contain some interruptions such as moving cars, pedestrians, flying birds, and image alignment errors (see Fig.~\ref{fig_3}(e)). These interruptions share similar motion features but different appearance features with MAV, we can use an appearance-based classifier to classify them into MAV and clutter. Considering that the appearance-based classifier will be called frequently and most of the common CNN architectures such as ResNet\cite{resnet} and DenseNet\cite{densenet} have a very deep structure that requires large computation resources, a shallow CNN network (see Fig.~\ref{fig_6}) is used to extract the feature of MAV and classify candidate moving objects into MAV and clutter. We train the proposed CNN model with local images of candidate moving objects (details about the classification dataset are introduced in Section~\ref{dataset}). Each convolution layer uses ReLU activation and max pooling. The final layer uses a fully connected network with a softmax activation to classify the candidate moving objects into two classes. Up to now, we can obtain the final moving MAV, as shown in Fig.~\ref{fig_3}(f).
	
	\section{Experiment}\label{experiment}
	
	\subsection{Datasets}\label{dataset}
	To evaluate the performance of the proposed GLAD algorithm, we tested our proposed algorithm on three challenging datasets. Each dataset is briefly introduced below.
	
	\subsubsection{NPS-Drones dataset\cite{2021Fast}}
		This dataset contains 50 videos of custom delta wing air-frame with a total number of frames adding up to 70,250. Videos are captured by a GoPro 3 camera mounted on a custom delta-wing air-frame at HD resolution of 1920$\times$1280 or 1280$\times$960. Some sample frames are shown in Fig.~\ref{dataset_sample}(a). Objects in this dataset are mainly small drones ranging from 10$\times$8 to 65$\times$21. The average object size is 0.05$\%$ of the whole image size. We use the clean version annotations released by \cite{2021Dogfight}. Following the train/val/test split of \cite{2021Dogfight}, we use 40 videos for training and validation and 10 videos for testing.
	
	\subsubsection{Drone-vs-Bird dataset\cite{Drone-vs-Bird}}
\begin{figure}[h]
		\centering
		\subfloat[Sample frames from NPS-Drones dataset.]{\includegraphics[width=0.99\linewidth]{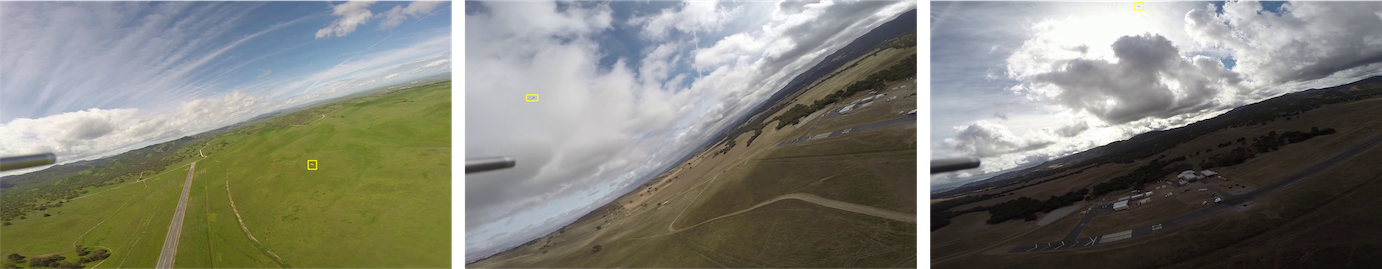}}\vspace{1mm}
		\subfloat[Sample frames from Drone-vs-Bird dataset.]{\includegraphics[width=0.99\linewidth]{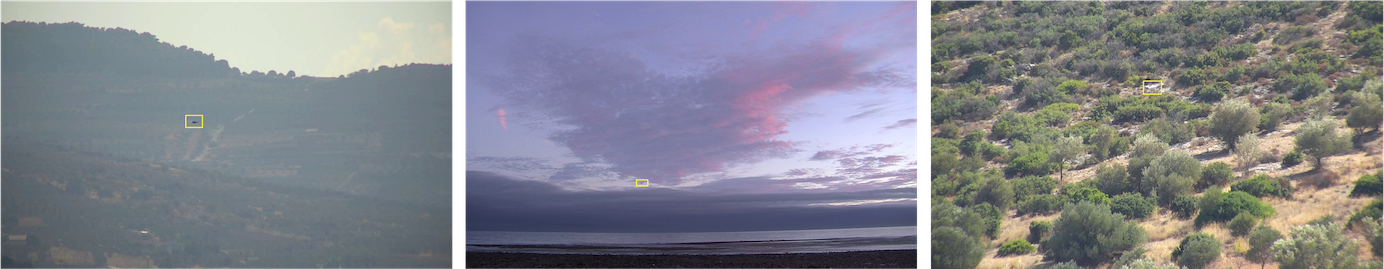}}
		\caption{Some sample frames from NPS-Drones dataset and Drone-vs-Bird dataset. \textbf{Yellow box} indicates the target drones.}
		\label{dataset_sample}
	\end{figure}

		This dataset is proposed in the International Workshop on Small-Drone Surveillance, Detection and Counteraction Techniques (WOSDETC), as part of the 16th IEEE International Conference on Advanced Video and Signal based Surveillance (AVSS). It contains 77 videos of 8 different types of drones with a total number of frames adding up to 104,760. Many of the videos are recorded with a static camera but also moving cameras are included. This dataset exhibits a high variability in difficulty, including sequences with sky or vegetation as background, different weather conditions, direct sun glare, and drastic camera movement. Moreover, objects in this dataset are mainly small drones captured at long distances and often surrounded by small interruptions, such as birds, and flying insects. Some sample frames are shown in Fig.~\ref{dataset_sample}(b). The average object size is 34$\times$23 (0.1$\%$ of the image size). Due to some errors and wrong annotations on two videos, we used 60 videos for training and validation and 15 videos for testing.
	
	\subsubsection{ARD-MAV dataset}
\begin{figure*}[!t]
		\centering
		\subfloat[Examples of original images and cropped images in ARD-MAV dataset. The cropped image is displayed at the top right of each image.]{\includegraphics[width=0.99\linewidth]{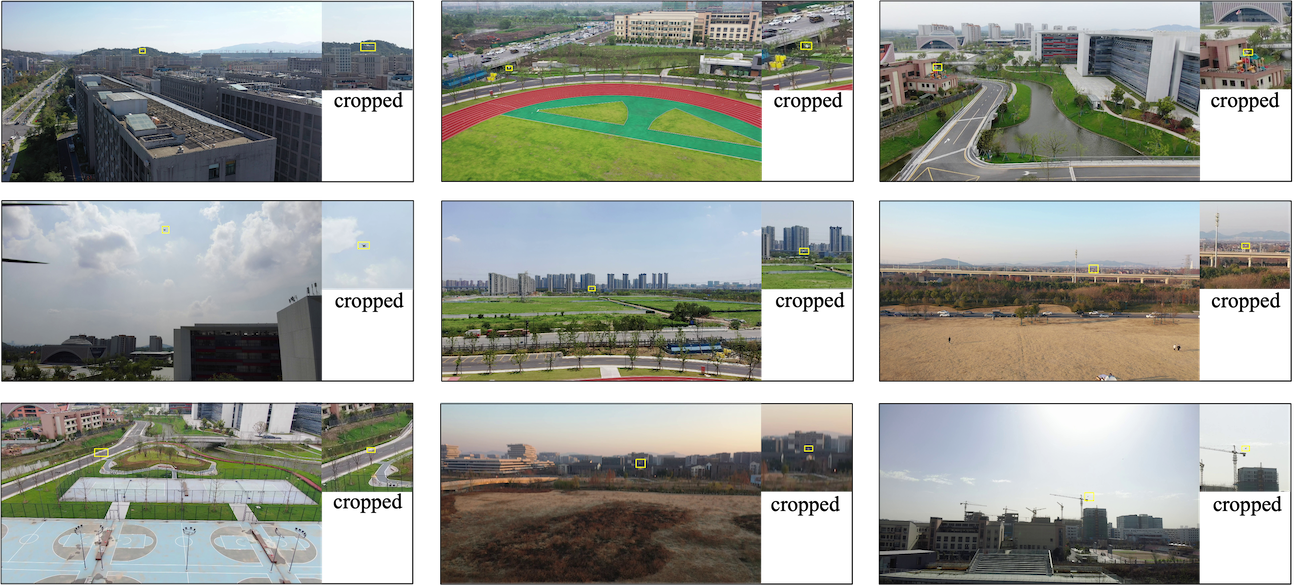}}\vspace{1mm}
		\subfloat[MAV's position distribution and size distribution in each part of ARD-MAV. The left picture is position distribution and the right is size distribution. ]{\includegraphics[width=0.61\linewidth]{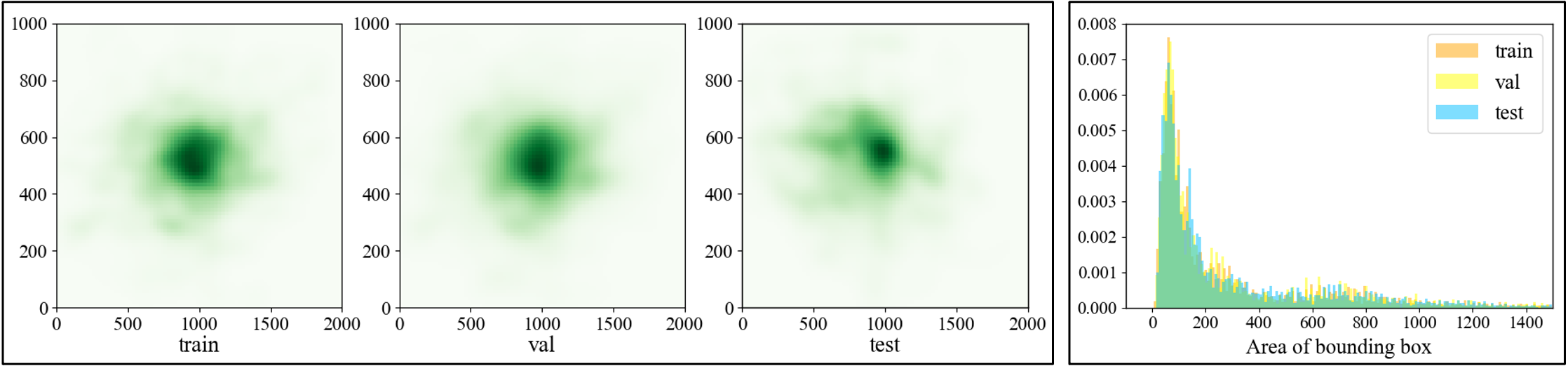}}\hspace{1mm}
		\subfloat[Examples of MAV and clutter. The left picture are examples of MAV, and the right picture are examples of clutter.]{\includegraphics[width=0.36\linewidth]{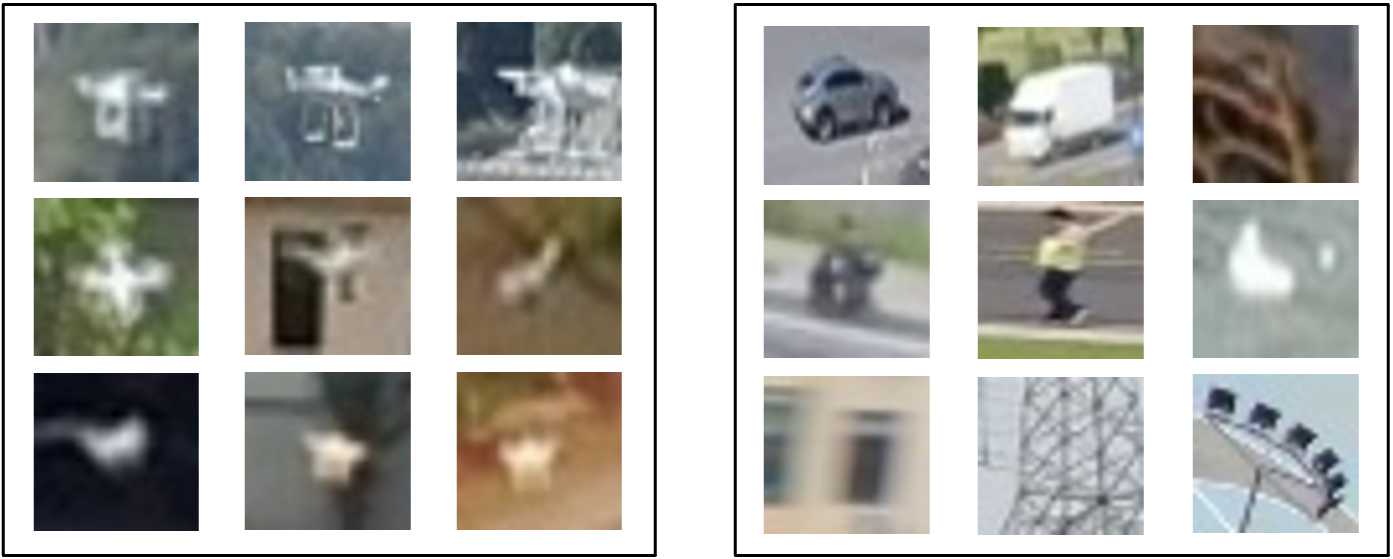}}
		\caption{Examples of our ARD-MAV dataset. The global appearance-based detection module is trained with original images. The local appearance-based detection module is trained with cropped images. The appearance-based classifier is trained with local images of MAV and clutter. \textbf{Yellow box} indicates the target MAVs.}
		\label{fig_7}
	\end{figure*}

	We created a dataset, named ARD-MAV, which contains 60 video sequences and 106,665 frames. All the videos are taken by the cameras of DJI Mavic2 Pro and M300 flying at low and medium altitudes. The videos are taken outdoor with different real-world challenges such as complex backgrounds, non-planar scenes, occlusion, abrupt camera movement, fast-moving MAVs, and small MAVs (some examples are shown in  Fig.~\ref{fig_7}(a)). Each video contains only one MAV and is about one minute long with a 30 FPS frame rate and a resolution of 1920$\times$1080. All the objects are manually labeled by labelImg. The object size ranges from 6$\times$3 to 136$\times$75. The smallest object refers to an MAV photoed by the aerial camera from more than 150m. The average object size is only 0.02$\%$ of the image size. As far as we know, this is the smallest average object size among the existing MAV datasets as shown in Table~\ref{tab1}.
	\begin{table}[t]
		\begin{center}
			\caption{Comparison of different MAV datasets}
			\label{tab1}
			\begin{tabular}{ccccc}
				\toprule 
				Dataset      & Count & Max Area & Min Area   & Average Area    \\
				\midrule 
				NPS-Drones      & 70250      & 6.6e-04   & 8.2e-05 & 0.05$\%$ \\
				FL-Drones      & 38948      & 1.4e-01   & 2.6e-04 & 0.07$\%$ \\
				DUT Anti-UAV     & 10109      & 7e-01  & 1.9e-06 & 1.3$\%$  \\
				Drone-vs-Bird    & 104760 & 2.5e-02 & 7.2e-06 & 0.1$\%$  \\
				ARD-MAV & 106665      & 3.5e-03  & 1.4e-05 & \textbf{0.02$\%$}  \\
				\bottomrule 
			\end{tabular}
		\end{center}
	\end{table}
	
	To train the global and local appearance-based detection module, we use 45 videos for training and validation, and the rest 15 videos for testing. The training images for the local appearance-based detection module are cropped around the center of the object with a size of 320$\times$320. Some examples of cropped images in the ARD-MAV dataset are shown in Fig.~\ref{fig_7}(a). Besides, the position distribution and size distribution are shown in Fig.~\ref{fig_7}(b).
	
	The training images for the local appearance-based classifier come from the intermediate results of moving object segmentation. The detection results after the motion-based classifier contain many non-MAV objects such as cars, pedestrians, swaying trees, shimmering water, and image alignment errors. Therefore, we manually divide these candidate moving objects into MAV and clutter (some examples are shown in Fig.~\ref{fig_7}(c)). The classification datasets include 46,268 images of clutter and 17,695 images of MAV. All the images are resized to 32$\times$32 before training and validation. 
	
	\subsection{Evaluation Metrics and Implementation Details}
	\subsubsection{Evaluation Metrics}
		Following the protocol in \cite{2021Dogfight}, the performance evaluation is based on Precision, Recall, F-Score, and AP. We set the intersection over union (IOU) threshold between predictions and ground truths to 0.5. Therefore, detected targets matching with ground truth with IOU > 0.5 are counted as true positives. In particular, the AP is calculated at 0.5 IOU threshold and is averaged over uniformly spaced 11 points of the precision-recall curve.
	
	\subsubsection{Implementation Details}
	Our evaluation experiments are implemented on a computer with an NVIDIA Geforce RTX 3070 GPU. For the training and validation of YOLOv5s, the input image size is down-sampled to 640$\times$640 as default. We use the Adam optimizer with a momentum of 0.937 and an initial learning rate of 0.01. We trained the model for 150 epochs with a batch size of 32. For the appearance-based classifier, the Adam optimizer is applied with a learning rate of 0.001. We trained the model for 100 epochs with a batch size of 64.
	
	To test the deployment efficiency of our approach on the mobile platform, we selected the NVIDIA Jetson Xavier NX processor for deployment experiments. This device has a 6-core CPU and 8G of GPU memory. We apply the TensorRT engine and pycuda API to accelerate the inference speed.
	
	Table~\ref{tab2} gives the parameters settings in this paper. These parameters are well-designed based on extensive experiments.
	
	\begin{table}[h]
		\begin{center}
			\caption{Table of parameters settings}
			\label{tab2}
			\begin{tabular}{ccccc}
				\toprule 
				Notation     & Description & Value \\
				\midrule 
				$T_0$      & Confidence threshold for global detector     & 0.5  \\
				$T_1$      & Confidence threshold for local detector     & 0.1  \\
				$T_2$      &  Threshold for frame difference binarization   & 5 \\
				$T_3$      &  Threshold for angle variance     & 0.8 \\
				$T_4$      &  Threshold for velocity variance     & 0.8 \\
				$T_5$      &  Threshold for velocity amplitude     & 1 \\
				$D_1$      &  Threshold for bounding box merging     & 15 \\
				\bottomrule 
			\end{tabular}
		\end{center}
	\end{table}
	
	\subsection{Evaluation Results under Different Conditions}
\begin{figure*}[]
		\centering
		\subfloat[Complex Backgrounds]{\includegraphics[width=0.325\linewidth]{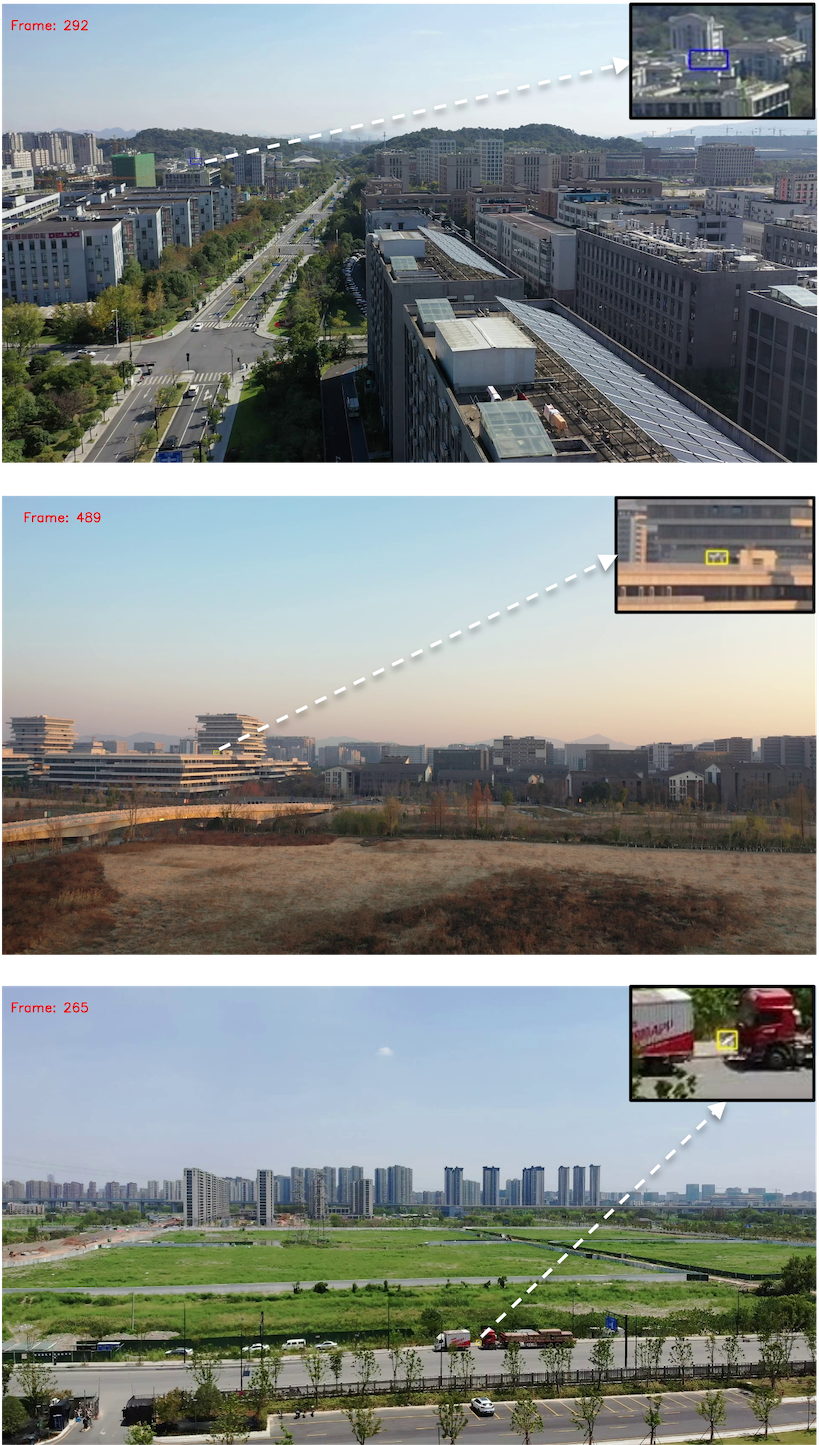}}\hspace{1mm}
		\subfloat[Small MAVs]{\includegraphics[width=0.325\linewidth]{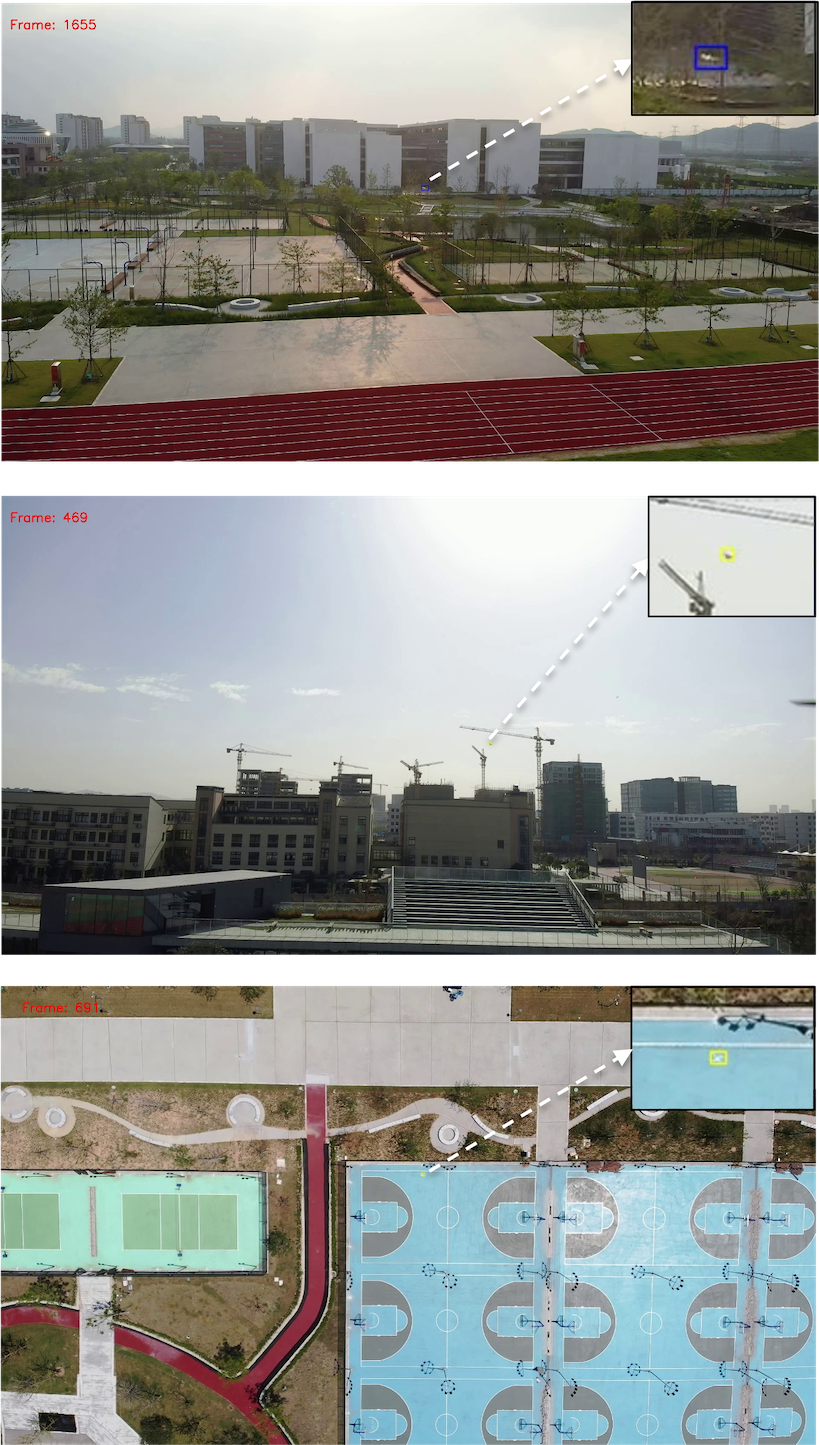}}\hspace{1mm}
		\subfloat[Ordinary Scenes]{\includegraphics[width=0.325\linewidth]{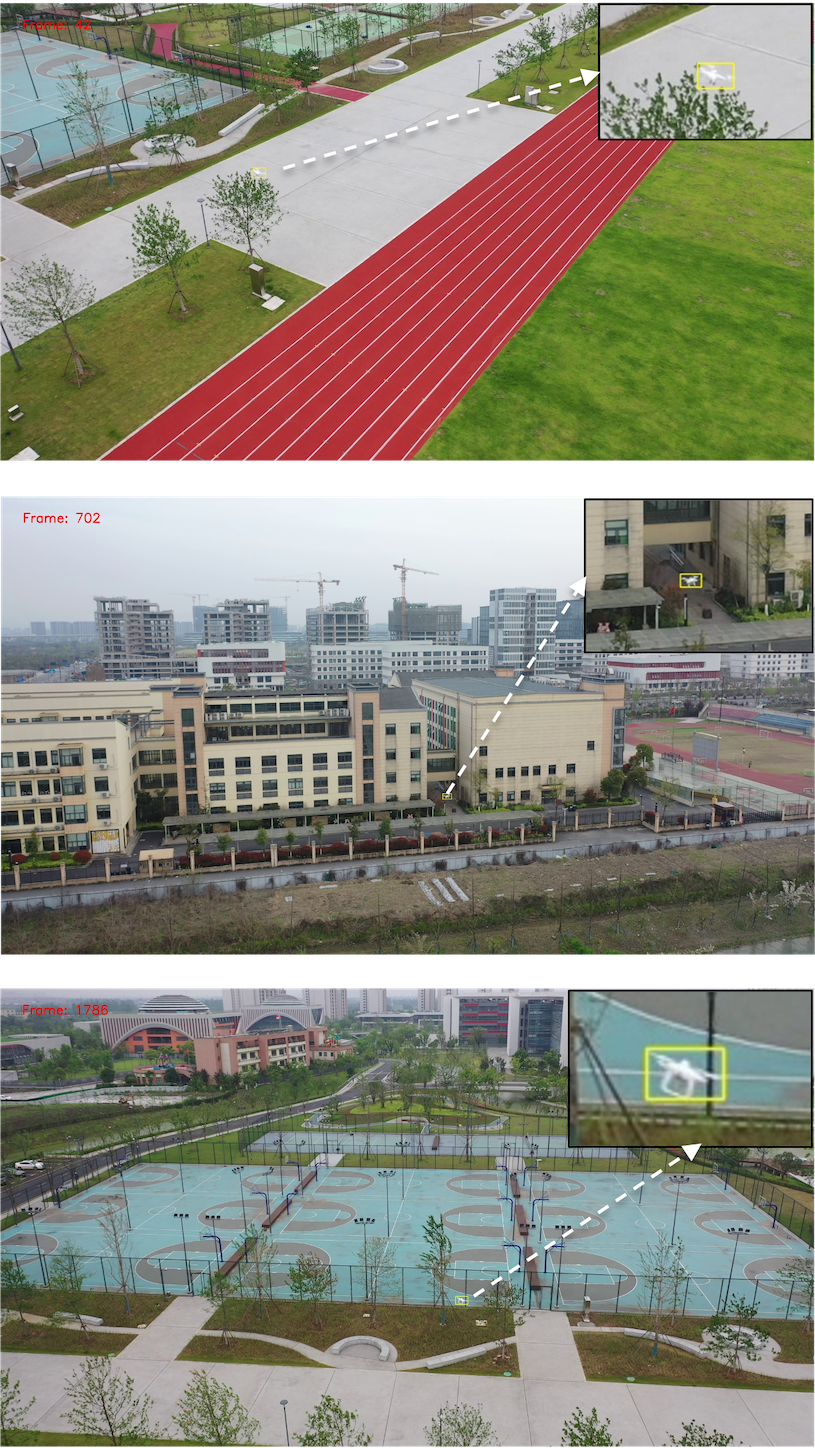}}
		\caption{Examples of detection results under various challenging conditions. \textbf{Blue box} indicates the target is detected by the motion-based detection module, \textbf{Yellow box} indicates the target is detected by the appearance-based detection module.}
		\label{fig_8}
	\end{figure*}

	We evaluate the proposed GLAD algorithm on 15 videos from the ARD-MAV dataset. According to the complexity of the background and target size, these videos are divided into ordinary scenes, complex backgrounds, and small MAVs. As shown in Table~\ref{tab3}, our proposed method can achieve a high success rate on ordinary scenes with nearly 100$\%$ precision and recall. For the cases of complex background and small MAVs, the detection accuracy degrades especially for small MAVs. It is important to note that the small MAV here denotes the MAV with a pixel area smaller than 100, which is difficult even for humans to recognize in the image. Some examples of detection results are illustrated in Fig.~\ref{fig_8}, we see that GLAD could successfully detect the MAV under challenging conditions.
	
	\begin{table}[]
		\begin{center}
			\caption{The detection results of our proposed GLAD algorithm on different conditions}
			\label{tab3}
			\begin{threeparttable}
				\begin{tabular}{ccccccc}
					\toprule 
					Conditions & Count & $S_{bbox}\tnote{1}$& Precision & Recall & F-Score  & AP\\
					\midrule 
					Ordinary & 9326 & 726  & 0.99 & 0.96 & 0.97 & 0.91 \\
					Complex & 9649 & 265  & 0.94 & 0.86 & 0.90 & 0.81 \\
					Small & 9347 & 63  & 0.82 & 0.67 & 0.73 & 0.58 \\
					Total & 28322 & 350 & 0.92 & 0.82 & 0.87 & 0.80\\
					\bottomrule 
				\end{tabular}
				\begin{tablenotes}
					\footnotesize
					\item[1] $S_{bbox}$: Average pixel area of bounding boxes.
				\end{tablenotes}
			\end{threeparttable}			
		\end{center}
	\end{table}
	
	\subsection{Comparison with Prior Works}\label{section_prior}
	We compare our proposed GLAD algorithm with several state-of-the-art methods on the ARD-MAV dataset. 
		Specifically, the YOLOv5s-1 and TPH-YOLOv5l-1 uses the default 640$\times$640 input image size, and YOLOv5s-2 and TPH-YOLOv5l-2 uses the 1536$\times$1536 input image size for inference. All compared methods are implemented based on official codes and are fine-tuned using the pre-trained weights available with public codes.
	
	The quantitative comparison of GLAD with the compared methods on the ARD-MAV dataset is shown in Table~\ref{tab4}. Our proposed algorithm outperforms the existing methods both on detection accuracy and computational efficiency. It is worth noting that GLAD significantly outperforms the existing methods on recall metric.
	
	The quantitative comparison of GLAD with other methods on the  NPS-Drones dataset and Drone-vs-Bird dataset are shown in Table~\ref{tab_NPS} and Table~\ref{tab_DvB} respectively. The experimental results of the compared methods are from \cite{2021Dogfight} and \cite{2021DT-Benchmark}. The results demonstrate that our proposed algorithm outperforms the existing methods on different evaluation metrics, especially on recall metric.
	
\begin{figure}[h]
		\centering
		\includegraphics[width=0.99\linewidth]{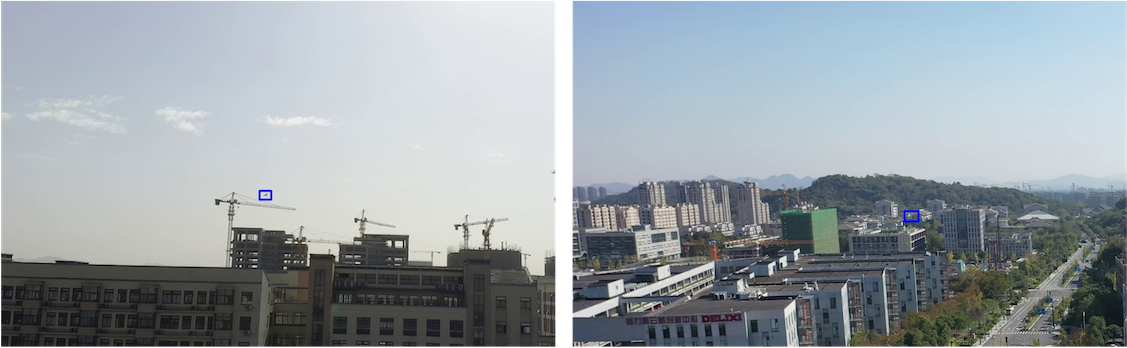}
		\caption{Some examples of the successful detection of GMD when GAD fails to detect a target.}
		\label{GMOD}
	\end{figure}

	We attribute the better performance of GLAD over compared methods to several factors. First, GLAD adopts a local search region that greatly retains the valuable appearance information of small targets and targets under complex backgrounds. However, the down-sampling method adopted in compared methods such as YOLOv5, TPH-YOLOv5, and MEGA all lead to huge information loss. The tremendous improvements on recall metric when enlarging the input image size of YOLOv5s and TPH-YOLOv5 exactly confirm it. Second, the motion-based detection module can make good use of the motion features and detect the target MAV when the appearance features are unreliable under challenging conditions (some examples are shown in Fig~\ref{GMOD}). Third, our proposed GLAD algorithm can detect stationary and moving targets simultaneously because we use the appearance features and motion features independently on each module. However, the compared methods such as  Dogfight\cite{2021Dogfight} try to detect MAV by jointly using appearance features and motion cues. As a result, the stationary target and slow-moving target are hard to detect.
	
	\begin{table}[]
		\begin{center}
			\caption{Quantitative comparison of the GLAD with state-of-the-art methods on ARD-MAV dataset}
			\label{tab4}
			\begin{tabular}{cccccc}
				\toprule 
				Method & Precision & Recall & F-Score & AP & FPS\\
				\midrule 
				YOLOv5s-1 & 0.90 & 0.20 & 0.33 & 0.56 & \textbf{149.3}\\
				YOLOv5s-2  & 0.78 & 0.41 & 0.54 & 0.61& 88.5\\
				TPH-YOLOv5l-1 \cite{TPH-YOLOv5} & 0.87 & 0.27 & 0.41 & 0.58 & 51.5\\
				TPH-YOLOv5l-2 \cite{TPH-YOLOv5} & 0.82 & 0.58 & 0.68 & 0.73 & 12.8\\
				Dogfight\cite{2021Dogfight} & 0.54 & 0.27 & 0.36 & 0.22 & 1.0\\
				MEGA\cite{mega} & 0.45 & 0.35 & 0.39 & 0.31 &3.5 \\
				GLAD & \textbf{0.92} & \textbf{0.82} & \textbf{0.87} & \textbf{0.80} & 146.5 \\
				\bottomrule 
			\end{tabular}
		\end{center}
	\end{table}
	
	\begin{table}[]
		\begin{center}
			\caption{Quantitative comparison of the GLAD with state-of-the-art methods on NPS-Drones dataset}
			\label{tab_NPS}
			\begin{tabular}{ccccc}
				\toprule 
				Method & Precision & Recall & F-Score & AP\\
				\midrule 
				SCRDet-H\cite{SCRDet} & 0.81 & 0.74 & 0.77 & 0.65\\
				SCRDet-R\cite{SCRDet}  & 0.79 & 0.71 & 0.75 & 0.61\\
				FCOS \cite{FCOS} & 0.88 & 0.84 & 0.86 & 0.83\\
				Mask-RCNN\cite{Mask-RCNN}  & 0.66 & 0.91 & 0.76 & \textbf{0.89}\\
				MEGA\cite{mega} & 0.88 & 0.82 & 0.85 & 0.83 \\
				SLSA\cite{SLSA} & 0.47 & 0.67 & 0.55 & 0.46 \\
				Dogfight\cite{2021Dogfight} & \textbf{0.92} & 0.91 & 0.92 & \textbf{0.89}\\
				GLAD & \textbf{0.92} & \textbf{0.95} & \textbf{0.93} & \textbf{0.89} \\
				\bottomrule 
			\end{tabular}
		\end{center}
	\end{table}
	
	\begin{table}[]
		\begin{center}
			\caption{Quantitative comparison of the GLAD with state-of-the-art methods on Drone-vs-Bird dataset}
			\label{tab_DvB}
			\begin{tabular}{cccccc}
				\toprule 
				Method & Faster-RCNN & SSD512 & YOLOv3 & DETR & GLAD \\
				\midrule 
				AP & 0.632 & 0.629 & 0.546 & 0.667 & \textbf{0.701}\\
				\bottomrule 
			\end{tabular}
		\end{center}
	\end{table}
	
	\subsection{Ablation Studies}
	In this section, we analyze different components of GLAD to verify their effectiveness. The experimental results in Table~\ref{tab5} show that each component of our proposed method is important and contributes to the final performance. 
	
	\begin{table}[]
		\begin{center}
			\caption{Ablation study of different components of our method}
			\label{tab5}
			\begin{threeparttable}
				\begin{tabular}{ccccc}
					\toprule 
					Methods & Precision & Recall & F-Score & AP \\
					\midrule 
					GAD & 0.76 & 0.17 & 0.28 & 0.18 \\
					GMD & 0.81 & 0.30 & 0.43 & 0.25 \\
					GAD+LAD & 0.90 & 0.51 & 0.65 & 0.54 \\
					GMD+LMD & 0.89 & 0.30 & 0.45 & 0.34 \\
					GAD+GMD+LAD & 0.90 & 0.78 & 0.84 & 0.72 \\
					GAD+GMD+LAD+LMD & 0.91 & 0.81 & 0.86 & 0.80\\
					GLAD & \textbf{0.92} & \textbf{0.82} & \textbf{0.87} & \textbf{0.80}\\
					\bottomrule 
				\end{tabular}
			\end{threeparttable}	
		\end{center}
	\end{table}
	
	\subsubsection{Influence of Motion-based Detection Module} The experimental results in Table~\ref{tab5} demonstrate that the motion-based detection module significantly improves detection accuracy, especially on recall metric compared with simple appearance-based methods. This improvement on recall metric might be due to the reason that the GMD can spot the subtle change between consecutive frames. Therefore, it can help generate an initial position for the local detector when the GAD fails to find an MAV under complex backgrounds and small MAV conditions. In addition, the LMD can also help the local detector maintain the right local search region when the LAD fails to detect a target.
	
	\subsubsection{Influence of Appearance-based Detection Module}
\begin{figure}[h]
		\centering
		\subfloat[Missed detection for hovering MAV]{\includegraphics[width=0.95\linewidth]{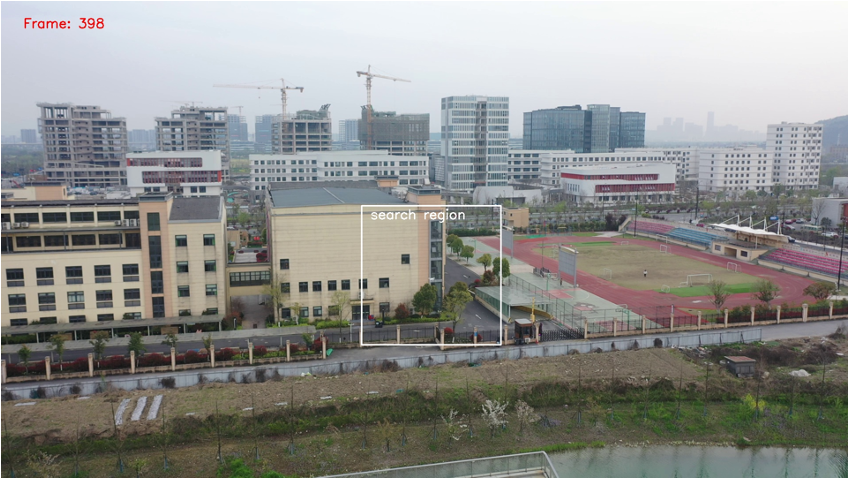}}\vspace{1mm}
		\subfloat[False detection for similar objects in local search region]{\includegraphics[width=0.95\linewidth]{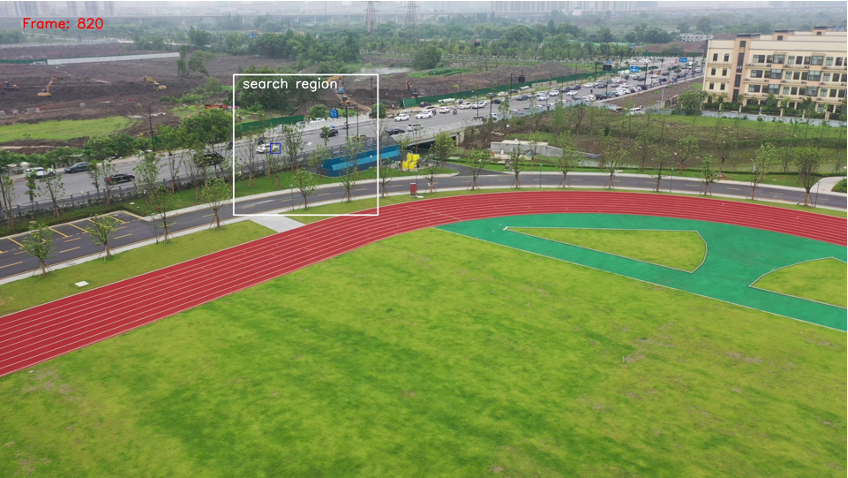}}
		\caption{Failure examples of the motion-based detection module.}
		\label{fig_10}
	\end{figure}

The experimental results in Table~\ref{tab5} show that the appearance-based detection module also plays an important role especially when there is a local search region compared with simple motion-based detection methods. On the one hand, this is because the motion-based module can only deal with moving targets. Hence, hovering and slow-moving MAVs are ignored. On the other hand, it owes to the higher robustness and reliability of the LAD than the LMD. In some cases, the LMD has difficulties in motion compensation or has low discernibility towards similar objects. Some failure examples of the local motion-based module are illustrated in Fig.~\ref{fig_10}.
	
	\subsubsection{Influence of Local Search Region}
	In theory, the local search region is conducive to improving the resolution of targets and removing interruptions. To verify the effectiveness of the local search region, we tested the performance of the appearance-based module and the motion-based module with and without a local search region. As shown in the first and third row of Table~\ref{tab5}, the local search region greatly improves the recall of the appearance-based module from 0.17 to 0.51. This huge improvement primarily owes to the higher resolution of the target in the local search region. When down-sampling the image from 1920$\times$1080 to 640$\times$640, the effective pixels of the target are dropped by nearly 80$\%$ and the appearance information is lost. In contrast, since the size of the local search region is smaller than the input size of YOLOv5s, the valuable appearance information of the target is completely reserved and the detection accuracy is greatly improved.
	
	On the other hand, the motion-based module benefits little from the local search region (as shown in the second and fourth row of Table~\ref{tab5}). The motion-based module primarily relies on the motion cues and local appearance features to locate and classify the MAV. However, the local appearance features in the local search region are the same as in the full-size image. Therefore, it just benefits from fewer interruptions in the local search region. In addition, a small patch occasionally impairs the motion-based module by inaccurate image alignment due to the sparse matched key points. As a result, the improvement of accuracy for the motion-based module is not as remarkable as the appearance-based module.
	
\begin{figure}[h]
		\centering
		\subfloat[The adaptive search region predicts a right target position when missing detection happens.]{\includegraphics[width=0.96\linewidth]{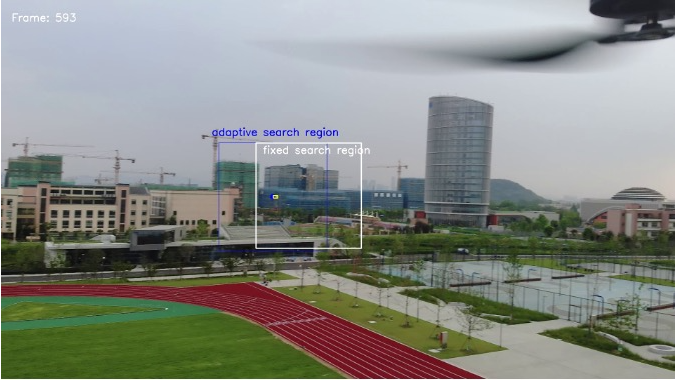}}\vspace{1mm}
		\subfloat[The adaptive search region has a better view than a fixed search region.]{\includegraphics[width=0.96\linewidth]{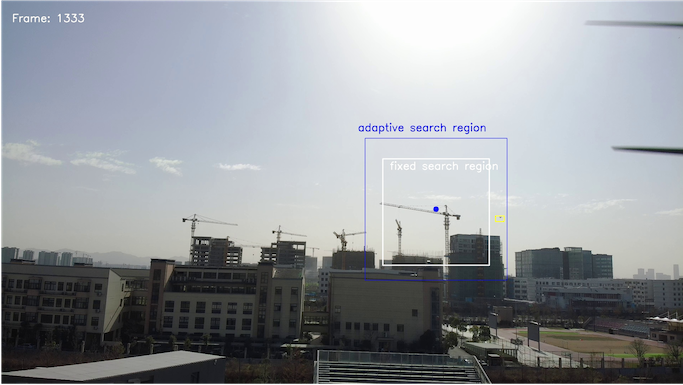}}
		\caption{Some examples to show the advantage of an adaptive search region. \textbf{Yellow box} indicates the position of the target MAV, \textbf{Blue box} indicates the adaptive search region, and \textbf{White box} indicates the fixed search region. The blue dot indicates the predicted target center.}
		\label{fig_asr}
	\end{figure}

	We have also tested the influence of the adaptive search region. As shown in the sixth and seventh row of Table~\ref{tab5}, the adaptive search region can improve the precision and recall. When occlusion and missing detection happen, a Kalman filter-based tracker can predict the target position in the coming frame, and an enlarged search region size has a better view than a fixed search region. Some examples are shown in Fig.~\ref{fig_asr}.
	
	\subsection{Inference Time}\label{inference}
	Table~\ref{tab7} demonstrates the frames per second (FPS) of each module on PC and Jetson Xavier NX respectively. We see that YOLOv5s is greatly accelerated with the TensorRT engine, it can reach 28.5 FPS on Jetson Xavier NX. The frame rate of the GMD and LMD is 10.6 FPS and 5.1 FPS on Jetson Xavier NX respectively. 
	
	Our method executes different modules based on the detection results of the previous frames rather than executing all of them sequentially. Since the detection difficulty varies across different videos, the inference time for each test video may differ. Therefore, we calculate the average FPS across all test videos to evaluate the performance of our proposed algorithm. The average running speed on PC and Jetson Xavier NX is 146.5 FPS and 23.6 FPS respectively. Although the motion-based detection modules cannot achieve real-time inference, they are called only when the GAD/LAD fails to detect a target under challenging conditions. Therefore, the average running speed can reach nearly real-time.
	\begin{table}[h]
		\begin{center}
			\caption{The FPS of each module on PC and Jetson Xavier NX}
			\label{tab7}
			\begin{threeparttable}
				\begin{tabular}{ccc}
					\toprule 
					Module & PC & Jetson Xavier NX\\
					\midrule 
					GAD/LAD & 149.3 & 28.5 \\
					GMD& 41.3 &  5.1 \\
					LMD  & 183.9 & 10.6 \\
					Average FPS on all test videos & 146.5  & 23.6 \\
					\bottomrule 
				\end{tabular}
		\end{threeparttable}	
	\end{center}
\end{table}

\section{Conclusion}\label{conclusion}
In this paper, we proposed a global-local MAV detector for air-to-air detection of MAVs under challenging conditions. The adaptive search region adopted in our approach significantly improved the detection accuracy by improving the resolution of the target using a small cropped image. Besides, we developed a motion-based detection module, which serves as a good assistant when the appearance features are unreliable. To evaluate the effectiveness of the proposed algorithm, we created a new dataset, named ARD-MAV, which contains various challenging conditions such as complex backgrounds, non-planar scenes, occlusion, abrupt camera movement, fast-moving MAVs, and small MAVs. Specifically, this dataset has the smallest average object size among current MAV detection datasets. Experiments on the three challenging datasets verified that the proposed algorithm can effectively detect MAV under various challenging conditions and outperforms state-of-the-art methods. Importantly, the experiments on the Jetson Xavier NX platform indicated that the proposed algorithm can be deployed on an aerial platform with real-time running speed.

In the future, our proposed algorithm will be extended to more types of MAVs. Moreover, an end-to-end network is necessary to be designed to simplify the training process, reduce the empirical parameters, and more effectively make use of the motion clues.

\bibliography{Reference_ghq} 

\begin{thebibliography}{10}

\bibitem{tang2018vision}
Y.~Tang, Y.~Hu, J.~Cui, F.~Liao, M.~Lao, F.~Lin, and R.~S. Teo, ``Vision-aided
  multi-{UAV} autonomous flocking in {GPS}-denied environment,'' {\em IEEE
  Transactions on Industrial Electronics}, vol.~66, no.~1, pp.~616--626, 2018.

\bibitem{2018TM}
R.~Opromolla, G.~Fasano, and D.~Accardo, ``A vision-based approach to {UAV}
  detection and tracking in cooperative applications,'' {\em Sensors}, vol.~18,
  no.~10, pp.~3391--3417, 2018.

\bibitem{2020marker}
M.~Vrba and M.~Saska, ``Marker-less micro aerial vehicle detection and
  localization using convolutional neural networks,'' {\em IEEE Robotics and
  Automation Letters}, vol.~5, no.~2, pp.~2459--2466, 2020.

\bibitem{Sapkota2016Vision}
K.~R. Sapkota, S.~Roelofsen, A.~Rozantsev, V.~Lepetit, D.~Gillet, P.~Fua, and
  A.~Martinoli, ``Vision-based unmanned aerial vehicle detection and tracking
  for sense and avoid systems,'' in {\em Proceedings of the 2016 IEEE/RSJ
  International Conference on Intelligent Robots and Systems (IROS)},
  pp.~1556--1561, 2016.

\bibitem{2021sense-avoid}
R.~Opromolla and G.~Fasano, ``Visual-based obstacle detection and tracking, and
  conflict detection for small {UAS} sense and avoid,'' {\em Aerospace Science
  and Technology}, vol.~119, pp.~107167--107186, 2021.

\bibitem{zhang2018survey}
J.~Zhang, K.~Zhang, J.~Wang, {\em et~al.}, ``A survey on anti-{UAV} technology
  and its future trend,'' {\em Advances in Aeronautical Science and
  Engineering}, vol.~9, no.~1, pp.~1--8, 2018.

\bibitem{2019Unlu}
E.~Unlu, E.~Zenou, N.~Riviere, and P.-E. Dupouy, ``Deep learning-based
  strategies for the detection and tracking of drones using several cameras,''
  {\em IPSJ Transactions on Computer Vision and Applications}, vol.~11, no.~1,
  pp.~1--13, 2019.

\bibitem{2020realworld}
M.~Pawe{\l}czyk and M.~Wojtyra, ``Real world object detection dataset for
  quadcopter unmanned aerial vehicle detection,'' {\em IEEE Access}, vol.~8,
  pp.~174394--174409, 2020.

\bibitem{2022Camera}
Y.~Zheng, C.~Zheng, X.~Zhang, F.~Chen, Z.~Chen, and S.~Zhao, ``Detection,
  localization, and tracking of multiple {MAV}s with panoramic stereo camera
  networks,'' {\em IEEE Transactions on Automation Science and Engineering},
  pp.~1--18, 2022.

\bibitem{Xie2021SmallLT}
J.~Xie, C.~Gao, J.~Wu, Z.~Shi, and J.~Chen, ``Small low-contrast target
  detection: Data-driven spatiotemporal feature fusion and implementation,''
  {\em IEEE Transactions on Cybernetics}, vol.~52, no.~11, pp.~11847--11858,
  2022.

\bibitem{2020DroneStaticCamera}
U.~Seidaliyeva, D.~Akhmetov, L.~Ilipbayeva, and E.~T. Matson, ``Real-time and
  accurate drone detection in a video with a static background,'' {\em
  Sensors}, vol.~20, no.~14, pp.~3856--3874, 2020.

\bibitem{2021DT-Benchmark}
B.~K.~S. Isaac-Medina, M.~Poyser, D.~Organisciak, C.~G. Willcocks, T.~P.
  Breckon, and H.~P.~H. Shum, ``Unmanned aerial vehicle visual detection and
  tracking using deep neural networks: A performance benchmark,'' in {\em
  Proceedings of the 2021 IEEE/CVF International Conference on Computer Vision
  Workshops (ICCVW)}, pp.~1223--1232, 2021.

\bibitem{2021Air}
Y.~Zheng, Z.~Chen, D.~Lv, Z.~Li, and S.~Zhao, ``Air-to-air visual detection of
  micro-{UAV}s: An experimental evaluation of deep learning,'' {\em IEEE
  Robotics and Automation Letters}, vol.~6, no.~2, pp.~1020--1027, 2021.

\bibitem{2022Anti-UAV-DT}
J.~Zhao, J.~Zhang, D.~Li, and D.~Wang, ``Vision-based anti-{UAV} detection and
  tracking,'' {\em IEEE Transactions on Intelligent Transportation Systems},
  pp.~1--12, 2022.

\bibitem{2021Fast}
J.~Li, D.~H. Ye, M.~Kolsch, J.~P. Wachs, and C.~A. Bouman, ``Fast and robust
  {UAV} to {UAV} detection and tracking from video,'' {\em IEEE Transactions on
  Emerging Topics in Computing}, vol.~10, no.~3, pp.~1519--1531, 2021.

\bibitem{NPU2020}
J.~Wang, G.~Zhang, K.~Zhang, Y.~Zhao, Q.~Wang, and X.~Li, ``Detection of small
  aerial object using random projection feature with region clustering,'' {\em
  IEEE Transactions on Cybernetics}, vol.~52, no.~5, pp.~3957--3970, 2020.

\bibitem{2017UDT}
L.~Du, C.~Gao, Q.~Feng, C.~Wang, and J.~Liu, ``Small {UAV} detection in videos
  from a single moving camera,'' in {\em Proceedings of the 2017 CCCV
  Communications in Computer and Information Science (CCIS)}, vol.~773,
  pp.~187--197, 2017.

\bibitem{wang2019flying}
C.~Wang, T.~Wang, E.~Wang, E.~Sun, and Z.~Luo, ``Flying small target detection
  for anti-{UAV} based on a gaussian mixture model in a compressive sensing
  domain,'' {\em Sensors}, vol.~19, no.~9, pp.~2168--2182, 2019.

\bibitem{Xie2020AdaptiveSS}
J.~Xie, J.~Yu, J.~Wu, Z.~Shi, and J.~Chen, ``Adaptive switching
  spatial-temporal fusion detection for remote flying drones,'' {\em IEEE
  Transactions on Vehicular Technology}, vol.~69, pp.~6964--6976, 2020.

\bibitem{Rozantsev2017DetectingFO}
A.~Rozantsev, V.~Lepetit, and P.~V. Fua, ``Detecting flying objects using a
  single moving camera,'' {\em IEEE Transactions on Pattern Analysis and
  Machine Intelligence}, vol.~39, pp.~879--892, 2017.

\bibitem{2021Dogfight}
M.~W. Ashraf, W.~Sultani, and M.~Shah, ``Dogfight: Detecting drones from drones
  videos,'' in {\em Proceedings of the 2021 IEEE/CVF Conference on Computer
  Vision and Pattern Recognition (CVPR)}, pp.~7063--7072, 2021.

\bibitem{wang2023RAFT}
H.~Wang, X.~Wang, C.~Zhou, W.~Meng, and Z.~Shi, ``Low in resolution, high in
  precision: {UAV} detection with super-resolution and motion information
  extraction,'' in {\em Proceedings of the IEEE International Conference on
  Acoustics, Speech and Signal Processing (ICASSP)}, pp.~1--5, IEEE, 2023.

\bibitem{2020hybridICUAS}
S.~Srigrarom and K.~H. Chew, ``Hybrid motion-based object detection for
  detecting and tracking of small and fast moving drones,'' in {\em Proceedings
  of the 2020 International Conference on Unmanned Aircraft Systems (ICUAS)},
  pp.~615--621, 2020.

\bibitem{zhang2021jointly}
P.~Zhang, J.~Zhao, C.~Bo, D.~Wang, H.~Lu, and X.~Yang, ``Jointly modeling
  motion and appearance cues for robust rgb-t tracking,'' {\em IEEE
  Transactions on Image Processing}, vol.~30, pp.~3335--3347, 2021.

\bibitem{mega}
Y.~Chen, Y.~Cao, H.~Hu, and L.~Wang, ``Memory enhanced global-local aggregation
  for video object detection,'' in {\em Proceedings of the 2020 IEEE/CVF
  Conference on Computer Vision and Pattern Recognition (CVPR)},
  pp.~10334--10343, 2020.

\bibitem{TPH-YOLOv5}
X.~Zhu, S.~Lyu, X.~Wang, and Q.~Zhao, ``{TPH}-{YOLO}v5: Improved {YOLO}v5 based
  on transformer prediction head for object detection on drone-captured
  scenarios,'' in {\em Proceedings of the 2021 IEEE/CVF International
  Conference on Computer Vision (ICCV) Workshops}, pp.~2778--2788, October
  2021.

\bibitem{Drone-vs-Bird}
F.~Coluccia, Angelo {\em et~al.}, ``Drone-vs-{B}ird detection challenge at
  {IEEE} {AVSS}2019,'' in {\em Proceedings of the 2019 IEEE International
  Conference on Advanced Video and Signal Based Surveillance (AVSS)}, pp.~1--7,
  2019.

\bibitem{2015Vision}
G.~Fatih, U.~Gokturk, S.~Erol, and K.~Sinan, ``Vision-based detection and
  distance estimation of micro unmanned aerial vehicles,'' {\em Sensors},
  vol.~15, no.~9, pp.~23805--23846, 2015.

\bibitem{2018Unlu}
E.~Unlu, E.~Zenou, and N.~Riviere, ``Using shape descriptors for {UAV}
  detection,'' {\em Electronic Imaging}, no.~9, pp.~1--5, 2018.

\bibitem{2021PruneYOLOv4}
H.~Liu, K.~Fan, Q.~Ouyang, and N.~Li, ``Real-time small drones detection based
  on pruned {YOLO}v4,'' {\em Sensors}, vol.~21, no.~10, pp.~3374--3390, 2021.

\bibitem{rui2021comprehensive}
C.~Rui, G.~Youwei, Z.~Huafei, and J.~Hongyu, ``A comprehensive approach for
  {UAV} small object detection with simulation-based transfer learning and
  adaptive fusion,'' {\em arXiv preprint arXiv:2109.01800}, 2021.

\bibitem{2017DeepCross-domain}
A.~Schumann, L.~Sommer, J.~Klatte, T.~Schuchert, and J.~Beyerer, ``Deep
  cross-domain flying object classification for robust {UAV} detection,'' in
  {\em Proceedings of the 14th IEEE International Conference on Advanced Video
  and Signal Based Surveillance (AVSS)}, 2017.

\bibitem{Li2016MultitargetDA}
J.~Li, D.~H. Ye, T.~Chung, M.~Kolsch, J.~Wachs, and C.~Bouman, ``Multi-target
  detection and tracking from a single camera in unmanned aerial vehicles
  ({UAV}s),'' in {\em Proceedings of the 2016 IEEE/RSJ International Conference
  on Intelligent Robots and Systems (IROS)}, pp.~4992--4997, 2016.

\bibitem{2015flying}
A.~Rozantsev, V.~Lepetit, and P.~Fua, ``Flying objects detection from a single
  moving camera,'' in {\em Proceedings of the 2015 IEEE/CVF Conference on
  Computer Vision and Pattern Recognition (CVPR)}, pp.~4128--4136, 2015.

\bibitem{2018effective}
S.~Minaeian, J.~Liu, and Y.-J. Son, ``Effective and efficient detection of
  moving targets from a {UAV}'s camera,'' {\em IEEE Transactions on Intelligent
  Transportation Systems}, vol.~19, no.~2, pp.~497--506, 2018.

\bibitem{2021uavdataset}
I.~Delibasoglu, ``{UAV} images dataset for moving object detection from moving
  cameras,'' {\em arXiv preprint arXiv:2103.11460}, 2021.

\bibitem{PLK}
J.-Y. Bouguet {\em et~al.}, ``Pyramidal implementation of the affine lucas
  kanade feature tracker description of the algorithm,'' {\em Intel
  corporation}, vol.~5, no.~1-10, p.~4, 2001.

\bibitem{resnet}
K.~He, X.~Zhang, S.~Ren, and J.~Sun, ``Deep residual learning for image
  recognition,'' in {\em Proceedings of the 2016 IEEE/CVF Conference on
  Computer Vision and Pattern Recognition (CVPR)}, pp.~770--778, 2016.

\bibitem{densenet}
G.~Huang, Z.~Liu, L.~Van Der~Maaten, and K.~Q. Weinberger, ``Densely connected
  convolutional networks,'' in {\em Proceedings of the 2017 IEEE/CVF Conference
  on Computer Vision and Pattern Recognition (CVPR)}, pp.~2261--2269, 2017.

\bibitem{SCRDet}
X.~Yang, J.~Yang, J.~Yan, Y.~Zhang, T.~Zhang, Z.~Guo, X.~Sun, and K.~Fu,
  ``{SCRD}et: Towards more robust detection for small, cluttered and rotated
  objects,'' in {\em Proceedings of the 2019 IEEE/CVF International Conference
  on Computer Vision (ICCV)}, pp.~8231--8240, 2019.

\bibitem{FCOS}
Z.~Tian, C.~Shen, H.~Chen, and T.~He, ``{FCOS}: Fully convolutional one-stage
  object detection,'' in {\em Proceedings of the 2019 IEEE/CVF International
  Conference on Computer Vision (ICCV)}, pp.~9626--9635, 2019.

\bibitem{Mask-RCNN}
K.~He, G.~Gkioxari, P.~Dollár, and R.~Girshick, ``Mask {R-CNN},'' in {\em
  Proceedings of the 2017 IEEE International Conference on Computer Vision
  (ICCV)}, pp.~2980--2988, 2017.

\bibitem{SLSA}
H.~Wu, Y.~Chen, N.~Wang, and Z.-X. Zhang, ``Sequence level semantics
  aggregation for video object detection,'' in {\em Proceedings of the 2019
  IEEE/CVF International Conference on Computer Vision (ICCV)}, pp.~9216--9224,
  2019.

\end{thebibliography}
\bibliographystyle{ieeetr}


\vfill

\end{document}